\documentclass[journal]{IEEEtran}
\usepackage{booktabs}
\usepackage{multirow}
\usepackage{amssymb}
\usepackage{diagbox}
\usepackage{graphicx}
\usepackage{amsmath}
\usepackage{algorithm}
\usepackage{algorithmic}
\usepackage{float}
\usepackage{threeparttable}
\ifCLASSINFOpdf
\else
\fi
\begin{document}
\title{Few-Shot Meta-Learning on Point Cloud \\for Semantic
Segmentation}

\author{Xudong~Li,
        Li~Feng,
        Lei~Li,
        Chen~Wang,
        and~Shuzhi~Sam~Ge,~\IEEEmembership{Fellow,~IEEE}% <-this % stops a space

% \thanks{Manuscript received March 29, 2021; revised xx xx, 2021. This work was supported by the National Natural Foundation of China U181320052.}% <-this % stops a space
\thanks{This work was supported by the National Natural Foundation of China under grant U181320052.}% <-this % stops a space
% \thanks{M. Shell was with the Department of Electrical and Computer Engineering, Georgia Institute of Technology, Atlanta, GA, 30332 USA e-mail: (see http://www.michaelshell.org/contact.html).}% <-this % stops a space
\thanks{Xudong Li, Li Feng, Lei Li, Chen Wang are with School of Computer Science and Engineering, University of Electronic Science and Technology of China,  Chengdu 611731, China(email:lixudong5211314@outlook.com; 1057265021@qq.com; 1534378592@qq.com; wangchen100@163.com).}% <-this % stops a space
\thanks{Shuzhi Sam Ge is with the Department of Electrical and Computer Engineering, National University of Singapore, Singapore 119077, Singapore(email:samge@uestc.edu.cn).}}
% email:samge@uestc.edu.cn

\markboth{Journal of \LaTeX\ Class Files,~Vol.~14, No.~8, April~2021}%
{Shell \MakeLowercase{\textit{et al.}}: Bare Demo of IEEEtran.cls for IEEE Journals}



\maketitle

\begin{abstract}

The promotion of construction robots can solve the problem of human resource shortage and improve the quality of decoration. To help the construction robots obtain environmental information, we need to use 3D point cloud, which is widely used in robotics, autonomous driving, and so on. With a good understanding of environmental information, construction robots can work better. However, the dynamic changes of 3D point cloud data may bring difficulties for construction robots to understand environmental information, such as when construction robots renovate houses. The paper proposes a semantic segmentation method for point cloud based on meta-learning. The method includes a basic learning module and a meta-learning module. The basic learning module is responsible for learning data features and evaluating the model, while the meta-learning module is responsible for updating the parameters of the model and improving the model generalization ability. In our work, we pioneered the method of producing datasets for meta-learning in 3D scenes, as well as demonstrated that the Model-Agnostic Meta-Learning (MAML) algorithm can be applied to process 3D point cloud data. At the same time, experiments show that our method can allow the model to be quickly applied to new environments with a few samples. Our method has important applications.

\end{abstract}

\begin{IEEEkeywords}
Artificial intelligence (AI), deep learning (DL), model-agnostic meta-learning (MAML), PointNet, point cloud semantic segmentation (PCSS).
\end{IEEEkeywords}

% \fi
\IEEEpeerreviewmaketitle


%for IEEE journal papers produced under \LaTeX\ using
%IEEEtran.cls version 1.8b and later.
% I wish you the best of success.

 

\section{Introduction}
\label{sec:introduction}
% 中国的人口老龄化在加速到来。人力资源短缺问题将越来越严重。装修机器人的发展可以解决房地产行业劳动力短缺问题，同时可以提高生产效率，保证生产安全。机器人在室内建筑装修环境中进行自动化施工，需要能够区分不同的施工对象，例如门、窗户、天花板、地板以及地板上的障碍物等。因此机器人需要对装修环境信息进行语义理解。机器人通常使用摄像头来获取环境信息。但是建筑工地通常光线不足，影响了摄像头拍摄的图片的质量。并且我们希望我们的机器人在晚上也能够工作。然而，激光雷达采集的三维点云数据不会因天气而受到影响。三维点云可以在空间中描述物体位置、形状、大小和其他属性，包含的信息比二位平面图像多，因此使用三维数据可以较为全面的获取周围环境的信息。同时，激光扫描仪、3D相机的普及，降低了获取三维数据的成本，降低了我们获取数据的难度。
\IEEEPARstart{C}{hina's} aging population is increasing significantly every day \cite{Rong_Chen2019201901213},\cite{Rong_Chen2018201701246}. And the problem of human resource shortage will create a huge need for robots. In real estate, construction robots can solve the labor shortage problem, while improving productivity and ensuring production safety. To automate the construction process, robots are required to recognize different objects, such as doors, windows, ceilings, floors, and so on. Or in other words, they are required to be able to semantically understand their surroundings. Normally, the robots obtain information from 2D cameras. However, the interior decoration scenes are poorly lit and quite gloomy, which affects the quality of the pictures taken by the camera. And robots also have to work at night. In contrast, 3D point cloud collected by LiDAR can not be affected by the weather. Moreover, the point cloud can describe the location, shape, size, and other attributes of objects, therefore it can provide more accurate geometric information than 2D images. Again, the popularity of laser scanners and 3D cameras has reduced the cost of acquiring 3D data and the difficulty of obtaining our data.

% 本文将重点研究3D点云的语义分割。PCSS将为每一个点生成语义信息。3D点云的语义分割可以分为传统的点云分割方法和以深度学习为基础的点云分割方法。
%The PCSS techniques will generate semantic information for every point. According to the review of current research results \cite{article_review}, PCSS is usually realized in two ways: regular supervised machine learning and deep learning.
% 机器监督学习的点云分割方法可以分为两类：个体PCSS和统计上下文PCSS。个体PCSS通过每个点的特征对点进行聚类分析。例如支持向量机、随机森\usepackage{}林等。个体分类器只考虑每个点的特征，而不考虑和周围点的共同特征，因此会出现一些噪声点导致结果不准确。统计上下文模型则考虑了每个点与周围点的关系。例如马尔科夫模型和条件随机场等。
% 基于深度学习的PCSS方法可以分为三类：分别是基于多视图的、基于体素的和基于点的。
% H. Su等人提出多视角的思想。将多个角度的二维数据输入到CNN中进行处理，之后将结果还原成三维，从而完成分类。
% D. Maturana等人将体素和3D CNNs结合，提出基于体素的三维卷积神经网络。在点云数据上创建三维立体的网格，通过体素数据在立方体空间上的相对位置信息来表示三维信息。
% 2017年，斯坦福大学的学者C. R. Qi等人[i]开创性的提出了PointNet网络，该网络直接将原始的三维点云数据作为输入应用于深度学习。PointNet在ModelNet40、ShapeNet数据集上均表现出最佳的分类结果。
In this paper, we specifically focus on 3D Point Cloud Semantic Segmentation (PCSS) techniques. PCSS is a major area of interest within the field of autonomous driving \cite{2018A},\cite{20203D}, virtual reality (VR) \cite{2016Progressive},\cite{2020PointXR}, and augmented reality (AR) \cite{2017Towards},\cite{2019Large}. However, PCSS is limited by the disadvantage of having large amounts of unstructured and unordered data \cite{app10072391}. Previous researchers have approached PCSS as a regular supervised machine learning problem, and their methods are mainly divided into two categories which are individual PCSS and statistical contextual models \cite{article_review},\cite{0Linking}. Individual PCSS performs cluster analysis of points by the features of each point, such as Support Vector Machines (SVM) \cite{article_svm_based},\cite{li2016three}, Random Forests \cite{chehata2009airborne}, AdaBoost \cite{Zhang2015A}, Bayesian Discriminant Classifiers \cite{2012Accuracy}, etc. Such approaches, however, have failed to address the influence of noisy points because they only consider the features of each point. The statistical contextual models, on the other hand, take the relationship between a single point and the surrounding points into consideration, such as Markov Networks \cite{inproceedings} and Conditional Random Fields \cite{niemeyer2014contextual}. They overcome the noise problem of initial labeling, and improve the accuracy of the segmentation.

Studies of Convolutional Neural Networks (CNNs) have greatly promoted the development of deep learning-based PCSS. However, CNNs can be adversely affected under certain conditions, such as the input data is unstructured point cloud data. To address this issue, some researchers have proposed various methods, which are mainly divided into three categories: multiview-based, voxel-based, and point-based \cite{0Linking}. H. Su et al. \cite{Su_2015_ICCV} proposed the Multiview-based PCSS method. They input 2D data from multiple perspectives into CNNs to extract features, and then transform the results to 3D to accomplish segmentation. Benefiting from the development of 3D CNNs, D. Maturana et al. \cite{maturana2015voxnet} combined voxels and 3D CNNs, and proposed Voxel-based 3D CNNs. They use a volumetric grid to represent the estimate of spatial occupancy and then input it to 3D CNNs to predict the class label directly. In 2017, C. R. Qi et al. \cite{qi2017pointnet}, scholars at Stanford University proposed the PointNet, which directly uses raw 3D point cloud data as input for deep learning applications. PointNet shows strong performance on both ModelNet40 \cite{wu20153d} and ShapeNet \cite{chang2015shapenet} datasets. The main shortcoming of PointNet is that the local features are ignored. To solve this issue,  C. R. Qi et al. \cite{qi2017pointnet++} proposed PointNet++, which splics a hierarchical neural network with PointNet to capture local representation within the data.

% 在我们的工作中，我们设计了一种直接处理点云数据、能够适应装修环境动态变化、有助于实现装修机器人自主性的点云语义分割方法。我们的方法分为两个模块：基础学习模conditions块和元优化模块。基础学习模块是为了学习支持集的特征。在基础学习模块，我们选择了PointNet作为基础学习器。采用N-way K-shot对S3DIS数据集中的数据进行采样，构建支持集和查询集。使用支持集对模型进行训练，使用查询集计算损失函数并对模型进行评估。元优化模块是为了提高模型的泛化能力。在元优化模块，我们通过损失函数值对模型的梯度进行更新，使得模型能够通过少量样本快速适应新的环境。
% We noticed that the modern PCSS methods require a large number of samples due to preventing overfitting. 
However, previous studies of PCSS have only focused on the accuracy under a large amount of data, they have not dealt with certain conditions in which we only have a small amount of labeled data. In this paper, we propose a new PCSS method which is able to directly process raw point cloud data and extract features from a small number of samples. Our network consists of two main modules. The first module is the basic learning module, which aims to extract the useful features of the support set constructed by N-way K-shot sampling method from the S3DIS dataset. Considering the extraordinary success and advantages of PointNet, we apply it as the base learner of the network. We used the support set to train the PointNet network and used the query set to calculate the loss value to evaluate the model. At the heart of our network is a meta-learning module, which aims to achieve finite-sample convergence and improve generalization ability. We update the gradient of the network by the loss values of the query set, which enables the model to quickly adapt to the new environment with a small number of samples. The main contributions of this paper are summarized as follows:
\begin{enumerate}
\item We propose a sampling method in order to build a 3D dataset with few samples for MAML algorithm.
\item We successfully apply the MAML algorithm over 3D point cloud to perform semantic segmentation.
\item We propose an effective network which can be rapidly adapted to a new environment with fewer samples, in order to significantly reduce the training time and the cost of data annotation.
\item We evaluate our model on a large-scale 3D indoor spaces dataset and compare it with a variety of state-of-the-art models.
\end{enumerate}

% 本文的主要贡献有以下：
% 1 数据集
% 2 将MAML算法成功从二维数据扩展到三维数据上，证明MAML算法能够应用于3D点云数据上。将MAML应用于点云语义分割任务中。
% 3 少样本学习 减少数据标注成本
% 4 结果
%\item We have successfully extended the MAML algorithm from 2D data to 3D data. We demonstrate that the MAML algorithm can be applied to 3D point cloud and can be used in point cloud semantic segmentation.
% \item We propose a model which can be quickly adapted to a new environment with fewer samples. It reduces the cost of data annotation for new environments. It can be used in various scenarios.
% \item We propose a model which has a higher accuracy rate and less training time. It has great application value.

% 全文分为五章。
The paper is composed of five themed sections and the remaining part is organized as follows. Section~\ref{sec:relate_works} presents the recent research on PointNet and Model-Agnostic Meta-Learning (MAML). Section~\ref{sec:methods} presents our network for point cloud semantic segmentation, focusing on the three key parts, which are N-way K-shot sampling method, the network architecture, and the algorithm. Section~\ref{sec:experiments} presents our experiments. We evaluated our model on the Stanford Large-Scale 3D Indoor Spaces Dataset (S3DIS) \cite{armeni20163d} and analyzed the accuracy of the results. Finally, the conclusion and future research directions are given in Section~\ref{sec:conclusion}.

% 为了让装修机器人能够动态适应环境变化，本文在神经网络模型中引入了少样本元学习方法，提出了一种基于元学习的3D点云语义分割方法。本文主要的贡献有以下三点：
%（1）室内装修环境数据集的预处理。本文通过使用 Faro Focus S70 三维激光扫描仪获取了装修环境数据集,并对其进行了冗余处理、滤波、精简和拼接等预处理,然后在预处理数据上进行标注。
%（2) 3D点云语义分割方法的改进。本文提出了基于 MAML 的点云语义分割算法,与传统的语义分割方法相比,该模型可以利用 MAML 算法学习 3D 点云数据任务集的共性特征,将其应用到新环境中,不仅可以使用少量训练数据拟合新环境,从而达到良好的分割效果,还可以提高模型训练效率。
%（3）点云语义分割方法应用于室内装修环境。将(2)中的技术方法应用于经过(1)处理的面向自主装修机器人的实际场景中,不仅验证了(1)中数据集处理方法的有效性,还验证了(2)中所提出的算法的可应用性。
% 

% \subsection{Subsection Heading Here}
% Subsection text here.
%\IEEEpubidadjcol
% \subsubsection{Subsubsection Heading Here}
% Subsubsection text here.

\section{Related Works}
\label{sec:relate_works}
% 在这个部分，我们从两个部分介绍相关工作：PointNet模型及相关技术和MAML模型。

% PointNet模型
\subsection{PointNet}
% 在第一部分提到，2017年，斯坦福大学Qi等人开创性地提出了PointNet网络。PointNet是第一个直接处理点云数据的神经网络模型，可用于点云的分类和语义分割任务。PointNet模型主要有三个关键模块：第一个模块是T-Net。T-Net网络能够预测出一个放射变换矩阵。将输入数据的矩阵与该矩阵相乘，使得点云数据在空间上对齐，方便之后的特征提取。第二个模块是最大池化模块。为了解决数据无序问题，PointNet使用了对称函数max-pooling来提取数据的全局特征。实验表明最大池化操作可以极大的增强网络的性能。第三个模块是局部特征与全局特征结合模块。最大池化操作后的数据是点云的全局信息。PointNet将全局信息与输入的点信息直接拼接在一起，通过卷积操作来提取特征，得到最终得到分类结果。PointNet在ModelNet40、ShapeNet数据集上均表现出最佳的分类效果，极大的推动了点云语义分割技术的发展。这些模块解决了点云无序性、置换不变形和旋转不变性等问题。在我们的工作中，我们将使用PointNet模型作为基础学习器。
% PointNet模型主要有三个关键模块，分别是处理点云特征的联合对齐网络T-Net、聚合点云特征信息的对称函数、局部特征与全局特征信息结合模块。
As it was mentioned in Section~\ref{sec:introduction}, in 2017, Qi et al. \cite{qi2017pointnet} at Stanford University proposed PointNet. PointNet is the first neural network that directly processes raw point cloud data for classification and semantic segmentation. It has three main key modules. The first module is T-Net. T-Net is a regression network. It can predict an input-dependent $3 \times 3$ transformation matrix. Multiplying the matrix of the input data with this matrix can make the point cloud data spatially aligned and facilitate subsequent feature extraction. The second module is max-pooling. To deal with unordered input data, PointNet uses the symmetric function max-pooling to extract the global features of the data. Experiments show that the max-pooling operation can greatly enhance the performance of the network. The third module is the combination module of local features and global features. PointNet concatenates the global feature vector with the input point vector directly and extracts the features through Convolutional Neural Network (CNN) to get the final classification and segmentation result. PointNet has shown the best segmentation results on both ModelNet40 \cite{wu20153d} and ShapeNet \cite{chang2015shapenet} datasets, and has greatly advanced the development of PCSS techniques. These modules solve the problems of disorder, substitution invariance and rotation invariance of point clouds.

% 然而，由于PointNet直接使用了最大池化来提取全局特征，因此其对局部特征提取不足。原作者提出了PointNet++网络。该网络借鉴了卷积神经网络中多层感受野的思想，通过不断分层来增大特征图的感受野，从而提高了每个数据包含的信息。PointNet++网络在点云的语义分割效果上有了一定的提升。
However, since PointNet directly uses max-pooling to extract global features, it is insufficient for local feature extraction. In a further study, Qi et al. \cite{qi2017pointnet++} proposed PointNet++. The network borrows the idea of multilayer perceptual field in CNNs, and captures local geometric features by using a hierarchical neural network. PointNet++ network has improved the semantic segmentation effect of point clouds.

% 
% Considering the extraordinary success and advantages of PointNet, we apply it as the base learner in our network.

% MAML模型
\subsection{Model-Agnostic Meta-Lerning}
% Meta-Learning，被称为元学习，是一种对模型的研究与学习。常见的深度学习模型主要是学习一个用于预测或分类的数学模型，而元学习学习的是学习的过程。因此元学习可以使用少量的训练样本来解决新的学习任务。MAML是与模型无关的元学习算法它通过优化参数在各个梯度方向的矢量和来对模型进行训练。模型的初始化参数为θ。MAML在分布任务t中随机选择若干个task进行采样，形成一个batch。之后对每个task分别计算损失函数值更新梯度值，得到θ1，θ2。在进行第一次梯度更新后，MAML可以进行第二次梯度更新。通过计算一个batch的损失总和，对梯度进行下降，更新θ。MAML可以在少样本的情况下快速适应新任务。MAML算法在二维图像监督学习和强化学习上得到了验证。Qian Zhong等人在MAML框架下，结合ResNet和GeM提取特征实现对遥感图片的检索。Zhang 等人将MAML算法与GAN网络结合，提出了MetaGAN网络，能够在少量样本情况下对图像进行分类。但是目前还没有相关文章证明MAML能够扩展到点云语义分割任务上。

% 而且，它适用于任何深度学习模型，可以解决分类、回归和强化学习等各种问题。
% 元学习的核心思想在于通过多个任务的学习对元模型进行优化，使其在面对新任务时可以进行快速而准确的特征学习。首先，通过多任务学习一个具有泛化性的模型。然后,在面对新任务时将该模型参数作为初始化参数,并在此基础上对使用少量数据对模型进行优化,从而达到快速适应新任务的目的。
Meta-Learning, which main purpose is to learn a rule. While common deep learning models focus on learning a mathematical model for prediction or classification, meta-learning learns the process of learning. Thus meta-learning can use a small number of training samples to solve a new learning task. Model-Agnostic Meta-Lerning (MAML) \cite{finn2017model} is a meta-learning algorithm that trains the model by optimizing the gradients of the parameters in each direction. The initialization parameter of the model is $\theta$. MAML randomly selects a number of tasks in task $\tau$ for sampling to form a batch. In the next step, the gradient values are updated by calculating the loss values for each task separately. After that, we get $\theta_{1},\theta_{2},\cdots,\theta_{n}$. After performing the first gradient update, MAML can perform a second gradient update. Finally, $\theta$ is updated by calculating the sum of losses of a batch. MAML can be quickly adapted to new tasks with few samples. The MAML algorithm has been validated for supervised machine learning and reinforcement learning on 2D images. Zhong et al. \cite{9191042} used the MAML framework as the basis and combined with ResNet and GeM to extract features to achieve the retrieval of remote sensing images. Zhang et al. \cite{NEURIPS2018_4e4e53aa} combined the MAML algorithm with GAN network and proposed MetaGAN network, which are capable of classifying images with a small number of samples. However, there are no relevant articles demonstrating that MAML can be applied to semantic segmentation of point clouds with few samples.

% Moreover, it is applicable to any deep learning model and can solve various problems such as classification, regression and reinforcement learning. 

\section{Methods}
\label{sec:methods}
% 基于MAML的3D点云语义分割算法

% 少样本元学习是通过少量样本学习到数据集的特征。尽管MAML是一种通用性较强的元学习的算法，但是目前还没有针对MAML算法的小样本点云数据集，也没有方法表明MAML在点云语义分割任务上的有效性。
% 因此，在这个部分，我们提出了基于点云的语义分割算法。
Few-shot meta-learning is to learn the features of the dataset by a small number of samples. Although MAML is a universal algorithm for meta-learning, there are no small samples of 3D point cloud dataset for the MAML algorithm and no methods which have demonstrated the effectiveness of MAML for 3D point cloud semantic segmentation. 

In this section, we propose a Few-Shot Meta-Learning Semantic Segmentation method for Point Cloud. In Section~\ref{sec:method_sampling}, we introduce the N-way K-shot sampling method to construct the training dataset and the test dataset. The dataset consisting of these data will be applied to the MAML algorithm. In Section~\ref{sec:method_arc}, we present the architecture of our model, including a basic learning module and a meta-learning module. Finally, we present the algorithm in Section~\ref{sec:method_alg}. We will demonstrate the effectiveness of MAML on 3D point cloud in the Section~\ref{sec:experiments}. 

% 使得分类器具有良好的泛化性能，避免出现过拟合问题。
%It enables the classifier to have good generalization performance and avoid overfitting problems. 

% N-way k-shot任务采样
\subsection{N-way K-shot sampling}
\label{sec:method_sampling}

% 在训练模型之前，我们需要构建训练集和测试集，我们分别称作支持集和查询集。我们将使用N-way K-shot方式进行采样，构成任务集，每个任务包括支持集和查询集。其中N表示样本的类别，K表示每个类别包含的样本数目。支持集包含N个标签数据。记为S=，表示D维的点云数据样本，表示数据对应的标签。查询集包含一些“未标记”的查询样本，记为Q。我们的模型在训练阶段的目的在于通过给定的支持集S和查询集Q预测出查询样本的正确标签。
Before training the model, we need to construct a training dataset and a test dataset, which we call the support set and the query set, respectively. We will use the N-way K-shot (N denotes the number of categories of samples and K denotes the number of samples contained in each category) method for sampling to form task sets, and each task set consists of a support set and a query set. The support set contains $N$ labeled data. There are $n$ categories with $k$ samples in each set, so $N = n \times k$. The support set is denoted as $S = \left\{(x_{1},y_{1}),(x_{2},y_{2}),\cdots,(x_{N},y_{N})\right\}$, $x_{i}\in R^{D}$ denotes the D-dimensional point cloud data, and $y_{i}\in \left\{1,\cdots,k\right\}$ denotes the corresponding label of the data and the semantic information of the data. The query set contains some "unlabeled" query samples, denoted as Q. The aim of our model in the training period is to predict the correct labels of the query samples from the given support set S and query set Q.

% 算法1为N-way K-shot采样方法。其中训练数据集D包含的数据类别为N，n表示任务集中的类别数量，k表示每类数据包含的数据样本数量，Ns表示支持集样本数量，Nq表示查询集样本数量，Rs函数表示从数据集Dtr中不放回的随机采取k个样本，D\S表示属于集合D且不属于集合S的所有元素。
% 在采样中，首先从训练集Dtr随机取n个类别，每个类别随机取(k+txk)个标注样本(t>=1)组成一个task，其中kxn个样本称为t的支持集，txkxn个样本称为t的查询集。我们需要不断采用算法1进行任务采样，生成训练任务集分布p(t)。
% The query set can be denoted as N-way K-shot task based on the number of categories and samples it contains. 
Algorithm \ref{alg:algorithm_N_way_K_shot} is the N-way K-shot sampling method. In the algorithm, the training dataset $D_{train}$ contains $N_{D}$ categories, $n$ denotes the number of categories in the task set, $k$ denotes the number of samples contained in each category, $N_{S}$ denotes the number of samples in support set, $N_{Q}$ denotes the number of samples in query set, the $Rs(D_{train},k)$ function denotes the random taking of $k$ samples from the dataset $D_{train}$ without putting back, and $D \backslash S$ denotes all samples belonging to the set D and not belonging to the set S. In sampling, $n$ categories are randomly taken from the training dataset $D_{train}$, and $(k+t \times k)$ labeled samples are randomly taken from each category to form a task $\tau$, where $k \times n$ samples are called the support set and $t \times k \times n$ samples are called the query set. We need to continuously employ Algorithm \ref{alg:algorithm_N_way_K_shot} to perform task sampling and generate the training dataset distribution $p(\tau)$. 

The output will be used as the dataset for the algorithm in Section~\ref{sec:method_alg}.

\begin{algorithm}
	\renewcommand{\algorithmicrequire}{\textbf{Input:}}
	\renewcommand{\algorithmicensure}{\textbf{Output:}}
	\caption{N-way K-shot sampling algorithm}
	\label{alg:algorithm_N_way_K_shot}
	\begin{algorithmic}[1]
		\REQUIRE $D_{train}$: training dataset;
		         $N_{D}$: data category;
		         $N_{S}$: number of support set samples, $N_{S} = n \times k$;
		         $N_{Q}$: number of query set samples, $N_{Q} = t \times n \times k$;
		\ENSURE $S$: support set; $Q$: query set
		\STATE $V = \left\{i, i = 1,\cdots,n | i \in N_{D}\right\}$
		\FOR{i in $\left\{1,\cdots,n\right\}$}
		\STATE $S_{i} = \left\{(x_{j},y_{j}), j = 1,\cdots,k | (x_{j},y_{j}) \in D_{train} \right\} \leftarrow Rs(D_{train}(V_{i}),k)$
		\STATE $Q_{i} \leftarrow Rs(D_{train}(V_{i}) \backslash S_{i}, t \times k)$
		\ENDFOR
		\STATE \textbf{return} $S=\left\{S_{i}\right\}, Q=\left\{Q_{i}\right\}$
	\end{algorithmic}  
\end{algorithm}

% 网络架构
\subsection{Architecture}
\label{sec:method_arc}
% 本文提出了一种基于MAML的3D点云语义分割网络模型。该模型开创性的将MAML应用于3D点云数据处理上。该模型分为两个部分：第一,首先将支持集数据作为模型的训练集对网络进行训练,然后将查询集数据作为模型的测试数据集进行测试;第二,利用测试数据的预测结果进行元优化操作，来更新模型的参数。为了更加清楚的表达整个处理流程,将整个网络分成了基础学习和元学习两个模块,网络设计如图所示。
In this paper, we propose a semantic segmentation network for point cloud based on meta-learning. The model is a pioneering application of MAML to the processing of 3D data. Our model can be divided into two parts. First, the support set is used as the training dataset of the model to train the network, and then the query set is used as the test dataset of the model for testing. Second, the test results are used to perform meta-learning operations to update the parameters of the model.

\begin{figure*}[htb]
\centering
\includegraphics[width=6.4in]{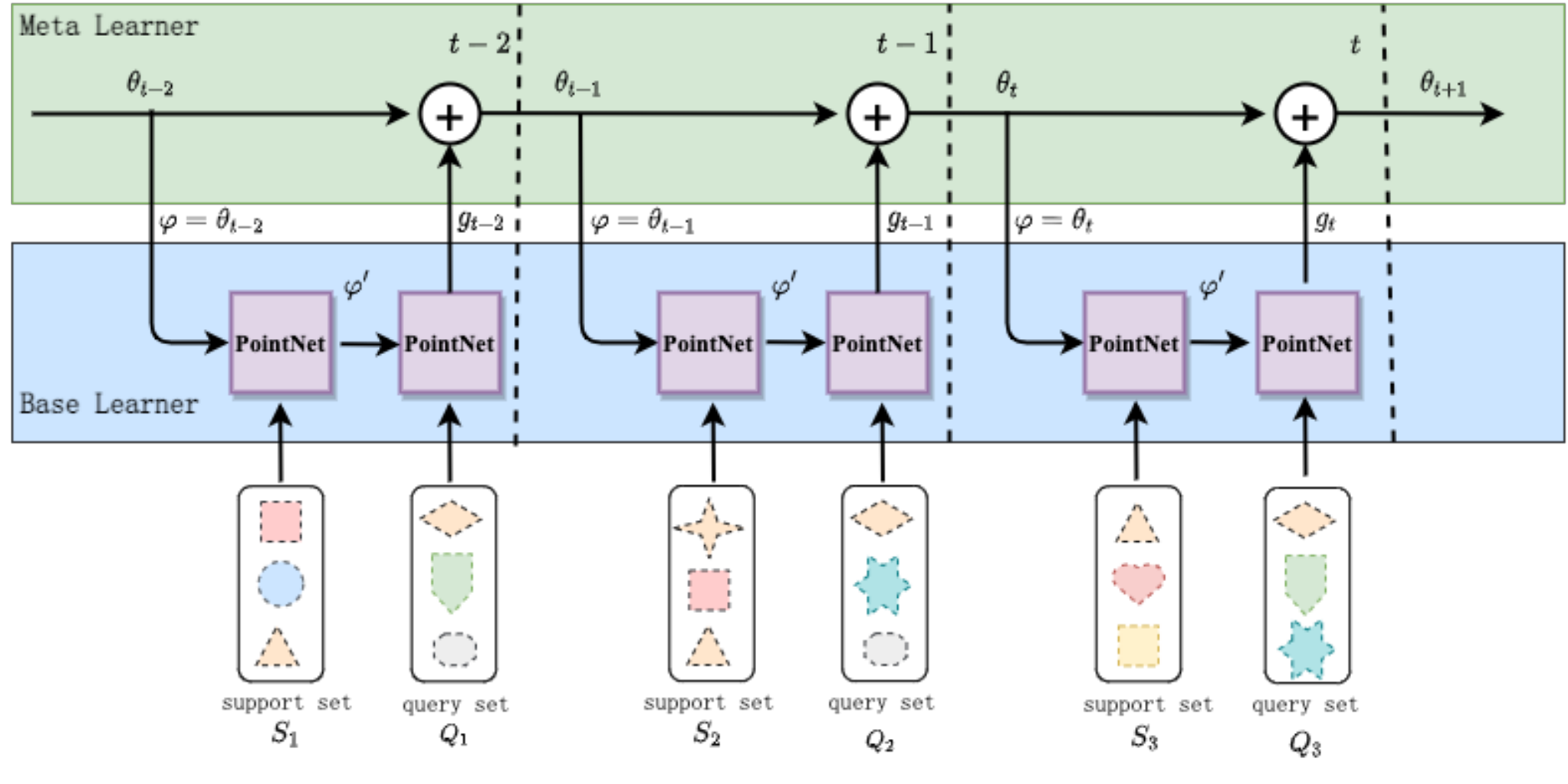}
\caption{Semantic segmentation network for point cloud based on meta-learning. The base learner takes the support set as input to train the model, and later takes the query set as input to evaluate the model. Meanwhile, the loss value is calculated. The meta learner will update the parameters of the model based on the loss value.}
\label{fig:MAML_3DPC_Arc}
\end{figure*}

To represent the model more clearly, the whole network is divided into two modules: basic learning module (base learner) and meta-learning module (meta learner). The network process flow is shown in Fig. \ref{fig:MAML_3DPC_Arc}.

% 基础学习模块
\subsubsection{Basic Learning Module}

% 基础学习模块,主要是通过PointNet模型学习支持集的特征。首先初始化模型的梯度参数，即 = 。然后对支持集数据训练得到支持集的损失值,如式1所示。
The basic learning module, which focuses on learning the features of dateset by PointNet. Firstly, the gradient parameter $\theta$ of the model is initialized, i.e. $\theta = \varphi$ . Then the loss values are obtained by training on the support set, as shown in (\ref{eq:BLM_Loss1}).

\begin{equation}
\begin{split}
\mathcal{L}_{S}(f_{\varphi})
& =\sum_{(x_{i},y_{i})\in S}^{} y_{i}\log_{}{f_{\varphi}{(x_{i})}} \\
& +(1-y_{i})log(1-\log_{}{f_{\varphi}{(x_{i}}}))
\label{eq:BLM_Loss1}
\end{split}
\end{equation}

% 计算出损失函数后,对梯度进行更新，如式1所示。
After the loss value is calculated, the gradient is updated as shown in (\ref{eq:BLM_Loss2}).

\begin{equation}
\begin{split}
{\varphi}^{'} & = \varphi - \beta \nabla_{\varphi}{\mathcal{L}_{S}(f_{\varphi})}
\\
& = \theta - \beta \nabla_{\theta}{\mathcal{L}_{S}(f_{\theta})}
\label{eq:BLM_Loss2}
\end{split}
\end{equation}

% 其中, β表示模型学习率。
where $\beta$ denotes the model learning rate.

% 最后我们利用查询集数据对PointNet模型进行评估，得到查询集的损失值，如式1所示。
Finally, we evaluate the PointNet model using the query set to obtain the loss values of the query set as shown in (\ref{eq:BLM_Loss3}).

\begin{equation}
\begin{split}
\mathcal{L}_{Q}(f_{{\varphi}^{'}})
& =\sum_{(x_{i},y_{i})\in Q}^{} y_{i}\log_{}{f_{{\varphi}^{'}}{(x_{i})}} \\
& +(1-y_{i})log(1-\log_{}{f_{{\varphi}^{'}}{(x_{i}}}))
\label{eq:BLM_Loss3}
\end{split}
\end{equation}

\begin{figure}[htb]
\centering
\includegraphics[width=3.2in]{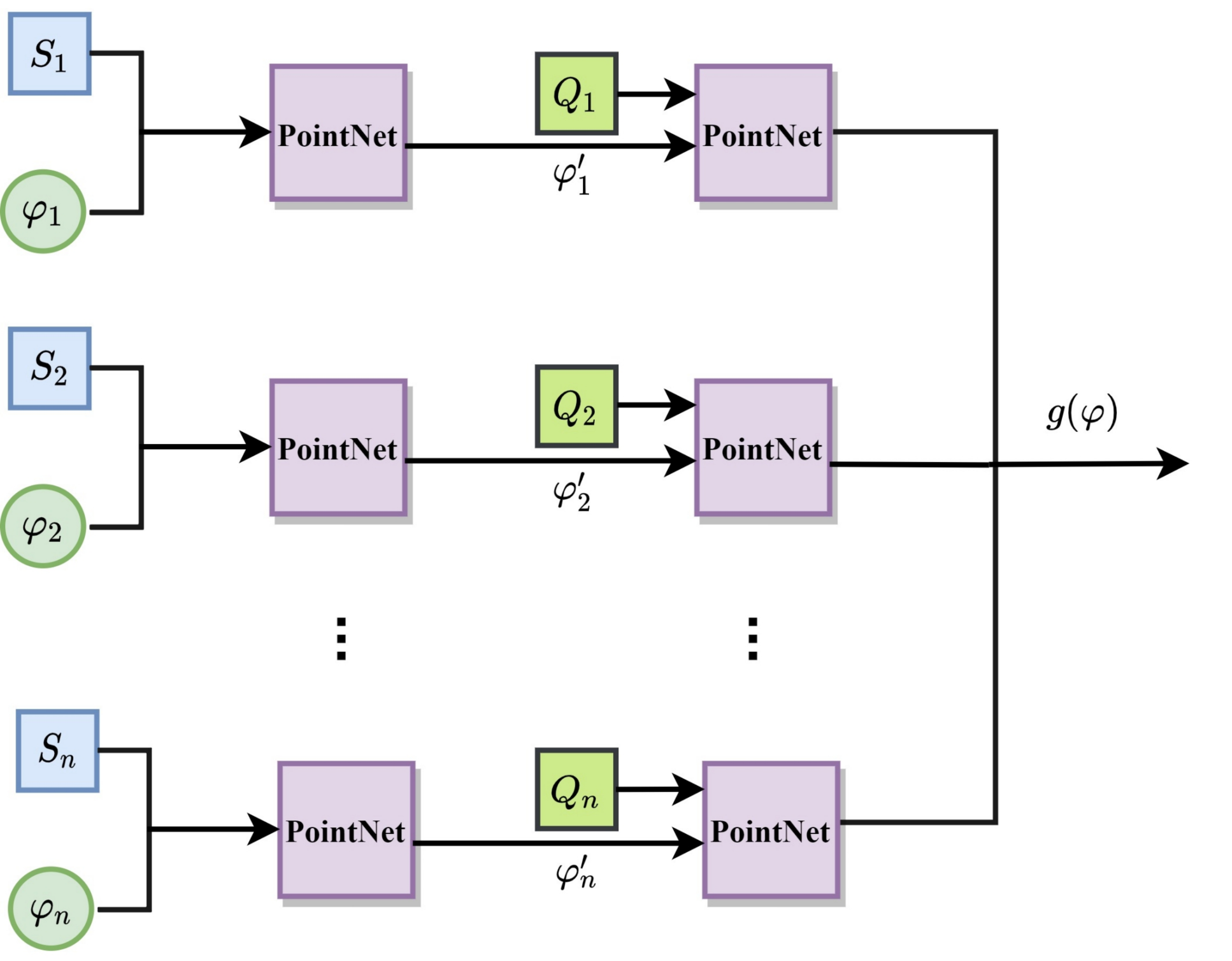}
\caption{Collaborative Network Model. At the beginning, the parameters $\varphi = \varphi_{1} = \varphi_{2} = \cdots = \varphi_{n}$. Then, the support sets $S_{1},S_{2},\cdots,S_{n}$ are fed into PointNet respectively, and the cross-entropy losses are calculated to update the gradient to obtain $\varphi_{1}^{'},\varphi_{2}^{'},\cdots,\varphi_{n}^{'}$. Next, take the query sets $Q_{1},Q_{2},\cdots,Q_{n}$ as input, calculate the average loss, and update $\varphi$ of the model.}
\label{fig:Collaborative_Network_Model}
\end{figure}

% 为了提高模型的效率，我们提出了协同网络模型，如图 所示。在该网络模型中，PointNet模型共享元学习器的初始化参数α，即α=α。然后利用支持集数据对模型进行训练，获取交叉熵损失函数，通过损失函数更新梯度，如式1和式1所示。最后分别使用查询集数据对模型进行评估，再计算所有网络的损失平均值，如式1所示。
To improve the efficiency of the model, we propose a collaborative network model, as shown in Fig. \ref{fig:Collaborative_Network_Model}. In this model, PointNet shares the initialization parameter $\varphi$, i.e., $\varphi = \varphi_{1} = \varphi_{2} = \cdots = \varphi_{n}$. Then the model is trained using the support set to obtain the cross-entropy loss, and the gradient is updated by the loss, as shown in (\ref{eq:BLM_Loss1}) and (\ref{eq:BLM_Loss2}). Finally, the model is evaluated using the query set separately and after that the loss values are averaged, as shown in (\ref{eq:BLM_Loss4}).

\begin{equation}
\begin{split}
\mathcal{L}_{Q}(f_{{\varphi}^{'}})
& = \frac{1}{n} \sum_{i=1}^{n} \mathcal{L}_{Q_{i}}(f_{{\varphi_{i}}^{'}}) \\
& = \frac{1}{n} \sum_{i=1}^{n} \mathcal{L}_{Q_{i}}(f_{\varphi_{i}-\beta \nabla_{\varphi_{i}}{\mathcal{L}_{S_{i}}(f_{\varphi})}}) 
\label{eq:BLM_Loss4}
\end{split}
\end{equation}

% 元优化模块
\subsubsection{Meta-Learning Module}

% 元优化模块主要是为了在少样本的情况下提高模型的泛化能力。在元优化模块中，我们通过所获取的损失函数值计算模型的梯度值并进行更新,如式 和 所示。
% 主要是为了查询集数据获得的损失函数值最小。为此,我们通过所获取的损失函数值计算模型的梯度值并进行更新,如式 和 所示。
The main purpose of the meta-learning module is to improve the generalization ability of the model with few samples. In this module, we calculate the gradient values of the model from the loss values and update them. This is shown in (\ref{eq:MLM_Loss1}) and (\ref{eq:MLM_Loss2}).

\begin{equation}
\begin{split}
g_{t}{(\nabla_{}{f(\theta_{t}),\varphi})} = - \alpha \nabla_{\theta_{t}}\mathcal{L}_{Q}(f_{\varphi-\beta \nabla _{\varphi}\mathcal{L}_{s}(f_{\varphi})})
\\
\label{eq:MLM_Loss1}
\end{split}
\end{equation}

%其中, 表示元优化步长。
where, $\alpha$ denotes the step in meta-learning.

\begin{equation}
\begin{split}
\theta_{t+1} = \theta_{t} + g_{t}{(\nabla_{}{f(\theta_{t}),\varphi})}
\label{eq:MLM_Loss2}
\end{split}
\end{equation}

%如果使用的是协同网络模型,则梯度值计算如式 所示。
If a collaborative network model is used, the gradient value is calculated as shown in (\ref{eq:MLM_Loss3}).

\begin{equation}
\begin{split}
g_{t}{(\nabla_{}{f(\theta_{t}),\varphi})}& = - \alpha \nabla_{\theta_{t}}\mathcal{L}_{Q}(f_{\varphi^{'}})
\\ & = \frac{1}{n} \sum_{i=1}^{n} \mathcal{L}_{Q_{i}}(f_{\varphi_{i}-\beta \nabla _{\varphi_{i}}\mathcal{L}_{s_{i}}(f_{\varphi})})
\\
\label{eq:MLM_Loss3}
\end{split}
\end{equation}

% 总体算法
\subsection{Algorithm}
\label{sec:method_alg}

% 算法1为本文提出的基于MAML的3D点云语义分割算法。
% 首先对模型进行随机初始化,然后从任务集中随机选择若干个任务组成一个 batch,针对这个 batch 的任务持续执行步骤5-8,最后循环执行直到模型收敛。

Algorithm \ref{alg:algorithm_MAML} is the semantic segmentation algorithm based on meta-learning proposed in this paper. The model parameters are first initialized randomly. Then a number of tasks are randomly selected from the task set to form a batch. Steps 5-8 are performed continuously for this batch of tasks until the model converges.

\begin{algorithm}
	\renewcommand{\algorithmicrequire}{\textbf{Input:}}
	\renewcommand{\algorithmicensure}{\textbf{Output:}}
	\caption{Semantic segmentation of point clouds based on meta-learning}
	\label{alg:algorithm_MAML}
	\begin{algorithmic}[1]
		\REQUIRE $D_{train}$: training dataset; $p(\tau)$: task set distribution; $\alpha$, $\beta$: learning rate 
		\ENSURE $\theta$
		\STATE Randomly initialize $\theta$
		\WHILE{not done}
		\STATE Randomly take a number of tasks from $p(\tau)$ to form a batch, suppose there are N $\tau$
		\FOR{i = 1 to N}
		\STATE Set $\varphi = \theta$
		\STATE $\varphi \leftarrow \varphi -\beta \nabla_{\varphi}\mathcal{L}_{s}(f_{\varphi})$
		\STATE Update $\mathcal{L}_{Q}(f_{\varphi})$ and $g{(\nabla_{}{f(\theta),\varphi})}$ based on Equation~(\ref{eq:BLM_Loss3}) and (\ref{eq:MLM_Loss1})
		\STATE $\theta \leftarrow \theta + g{(\nabla_{}{f(\theta),\varphi})}$
		\ENDFOR
		\ENDWHILE
		\STATE \textbf{return} $\theta$
	\end{algorithmic}  
\end{algorithm}

\section{Experiments}
\label{sec:experiments}

% 本小节的实验旨在验证以下三个问题：
% (1) MAML算法能否在3D点云数据集上应用成功。
% (2) 我们的算法是否会因为预训练数据集的大小不同而使得模型的效果不同？
% (3) 我们的算法能否使模型在少样本情况下快速应用于新任务。
% The experiments in this subsection aim to verify the following three questions:
% \begin{enumerate}
% \item Could the MAML algorithm successfully be applied on 3D point cloud dataset?
% \item Does our algorithm make the model perform differently due to the different sizes of pre-training dataset?
% \item Does our algorithm enable the model to be quickly applied to new tasks with few samples?
% \end{enumerate}
We first introduce the experiment environment in Section~\ref{sec:experiment_evironment} and the S3DIS dataset in Section~\ref{sec:experiment_S3DIS}. In Section~\ref{sec:experiment_exp}, we present our experiments. Experiments are divided into four parts. In Section~\ref{sec:experiment_exp1}, we verified that the MAML algorithm can be successfully applied to 3D point cloud and our model can be quickly adapted to new environments with few samples. In Section~\ref{sec:experiment_exp2}, we did cross-validation experiments and proved that our model has good generalisation ability. In Section~\ref{sec:experiment_exp3}, we visualised five samples to further demonstrate the validity of our proposed method. In Section~\ref{sec:experiment_exp4}, we have compared our method with other methods and have proven that our method has better results.

% 实验环境
\subsection{Experiment environment}
\label{sec:experiment_evironment}
%本论文实验在训练阶段主要是通过训练集数据进行训练,包括支持集和查询集,然后使用新任务训练数据集进行模型的适应训练,在测试阶段使用新任务的测试数据集进行测试,其中所有实验使用的平台环境如表 所示。
The experiment environment is shown in the Table \ref{tab:Exp_Env}.

\begin{table}[H]
\caption{Experiment environment}
\label{tab:Exp_Env}
\centering
\begin{tabular}{ll}
\toprule
Software and hardware & Model\\
\midrule
CPU     & Intel Core i7-8700 \\
GPU     & GeForce GTX 1080 Ti  \\
OS      & Ubuntu 18.04     \\
Computing Framework & TensorFlow-GPU 1.14 \\
GPU acceleration & cuda10.0/cuDNN7.4 \\
Programming language & Python2.7         \\
\bottomrule
\end{tabular}
\end{table}

% 数据集
\subsection{S3DIS dataset}
\label{sec:experiment_S3DIS}
% 我们将在S3DIS数据集上进行实验。S3DIS数据集是斯坦福大规模 3D 室内空间数据集。该数据集从3栋不同建筑楼采取，共计6个不同的大型场景，11种房间类型。如图为S3DIS数据集的六个区域。每个区域的室内房间类型个数如表所示。

\begin{figure}[htb]
\centering
\includegraphics[width=3.3in]{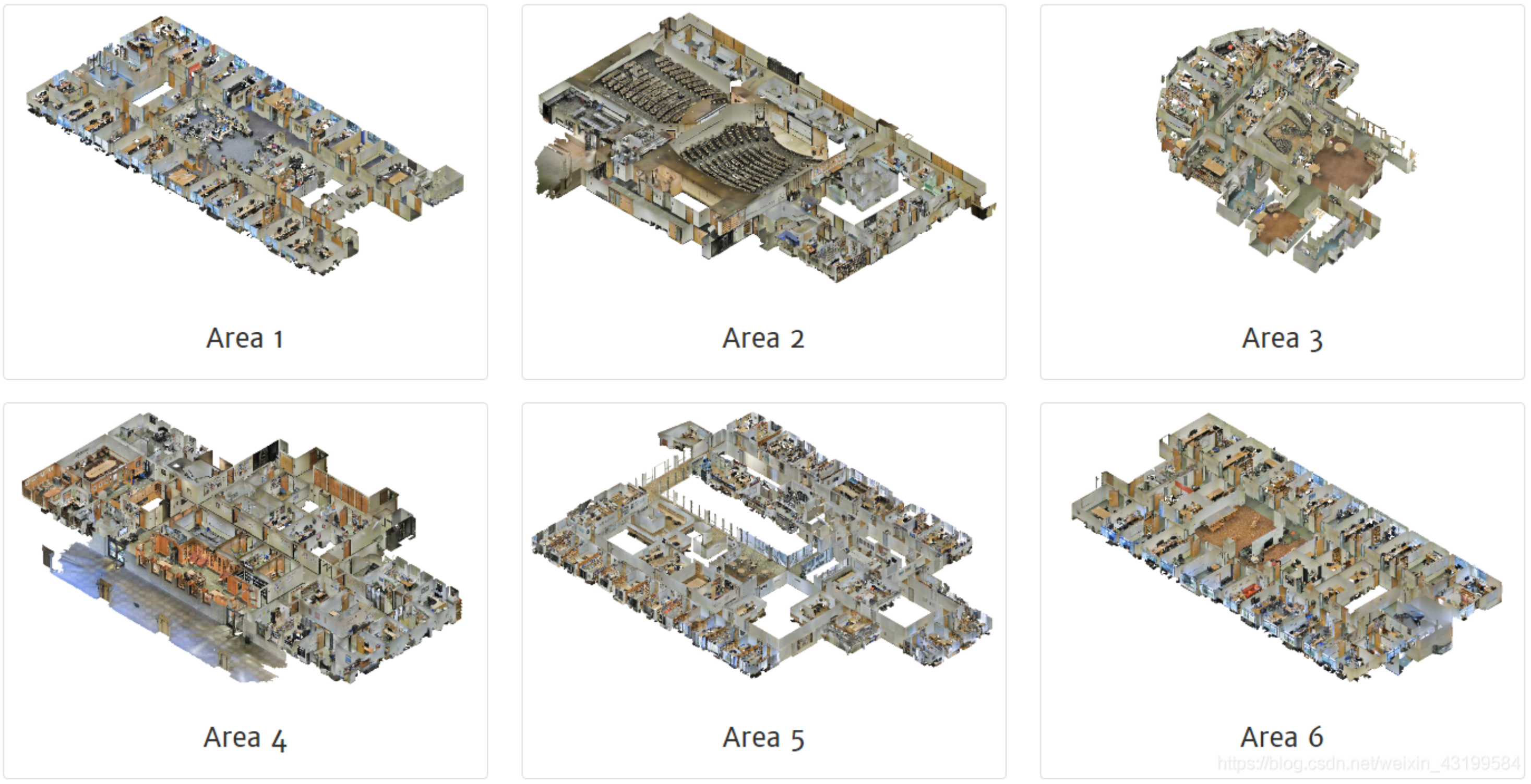}
\caption{S3DIS dataset. The dataset is composed of 6 large-scale indoor areas.}
\label{fig:S3DIS_Dataset}
\end{figure}

%如图所示。该数据集从3栋不同建筑楼采取，共计6个不同的大型场景，11种房间类型，分别为分别为办公室、礼堂、会议室、休息室、大堂、走廊、储藏室、食物储藏室、复印室、厕所和开放空间,每种类型又分为地板、天花板、横梁、墙壁、窗户、柱子、门、沙发、椅子、桌子、木板、书橱和其他杂物等 13 种语义类别,其中每个点云数据都包含坐标信息 XYZ、RGB 值深度信息和表面法向量等。
We will experiment on the Stanford Large-Scale 3D Indoor Spaces Dataset (S3DIS) \cite{armeni20163d}. The dataset is composed of 6 large-scale indoor areas from 3 different buildings, with a total of 11 room types \cite{armeni20163d}. The Fig. \ref{fig:S3DIS_Dataset} shows the six areas of the S3DIS dataset. The number of spaces in each area is shown in the Table \ref{tab:S3DIS_Dataset_Rooms}.

\begin{table*}[!htbp]
\caption{Disjoint spaces statistics per building area}
\label{tab:S3DIS_Dataset_Rooms}
\centering
\setlength{\tabcolsep}{1mm}{
\begin{tabular}{lcccccccccccc}
\toprule
~ & Office	& Conf. Room	& Auditorium	& Lobby	& Lounge	& Hallway	& Copy Room	& Pantry	& Open Space	& Storage	& WC	& Total Num \\
\midrule
Area 1	& 31  & 2	& -	& -	& -	& 8	 & 1	& 1	& -	& -	& 1	& 45 \\
Area 2	& 14  & 1	& 2	& -	& -	& 12 & -	& -	& -	& 9	& 2	& 39 \\
Area 3	& 10  & 1	& -	& -	& 2	& 6	 & -	& -	& -	& 2	& 2	& 24 \\
Area 4	& 22  & 3	& -	& 2	& -	& 14 & -	& -	& -	& 4	& 2	& 49 \\
Area 5	& 42  & 3	& -	& 1	& -	& 15 & -	& 1	& -	& 4	& 2	& 55 \\
Area 6	& 37  & 1	& -	& -	& 1	& 6	 & 1	& 1	& 1	& -	& -	& 53 \\
Total Num	& 156	& 11	& 2	& 3	& 3	& 61	& 2	& 3	& 1	& 19	& 9	& 270 \\
\bottomrule
\end{tabular}}
\end{table*}

% S3DIS数据集包含六个不同的区域，我们将整个数据集分为一个区域的预训练集和五个不同区域的训练测试集。例如Area1为预训练集，采用基于MAML迁移到Area2中，在迁移过程中我们随机采取 A的3D点云语义分割网络进行训练，然后将训练好的网络rea2的部分数据作为训练集对网络进行微调,然后使用其他数据进行测试。
% 室内装修环境 3D 点云数据集是使用 Faro Focus S70 激光扫描仪获取的真实点云数据,经过冗余处理、数据滤波、精简和拼接等四个步骤处理,形成一个大型室内场景,其中包括 14 个房间,主要包含地板、天花板、墙壁、柱子、横梁、窗户、门、走廊和其他杂物等语义类别,其中每个点云数据都包含坐标信息 XYZ 和颜色信息 RGB 值。

% 在数据准备过程中,我们首先将每个区域的点云数据按照房间进行分割,然后把每个房间按照 1× 1 的规格分割成块,每个块中包含很多个点云数据,每个点云数据包含点的空间坐标、RGB 颜色值和归一化坐标信息等 9 个维度的特征信息,并将其表示为
In the process of data preparation, we first partition the point cloud data of each area into rooms. Next, each room is divided into $1m \times 1m$ blocks, and each block contains lots of point cloud data. Each point contains 9 dimensions of feature information such as spatial coordinate, RGB color and normalized coordinate. And it is represented as $F=\left\{X,Y,Z,R,G,B,N_{X},N_{Y},N_{Z}\right\}$.

% S3DIS数据集语义分割
\subsection{Semantic segmentation of point clouds based on meta-learning}
\label{sec:experiment_exp}

% 学习率的选择
% \subsubsection{Learning Rate}

% Area1预训练数据集的语义分割
%\subsubsection{Transfer learning}

\begin{figure}[htb]
\centering
\includegraphics[width=3.3in]{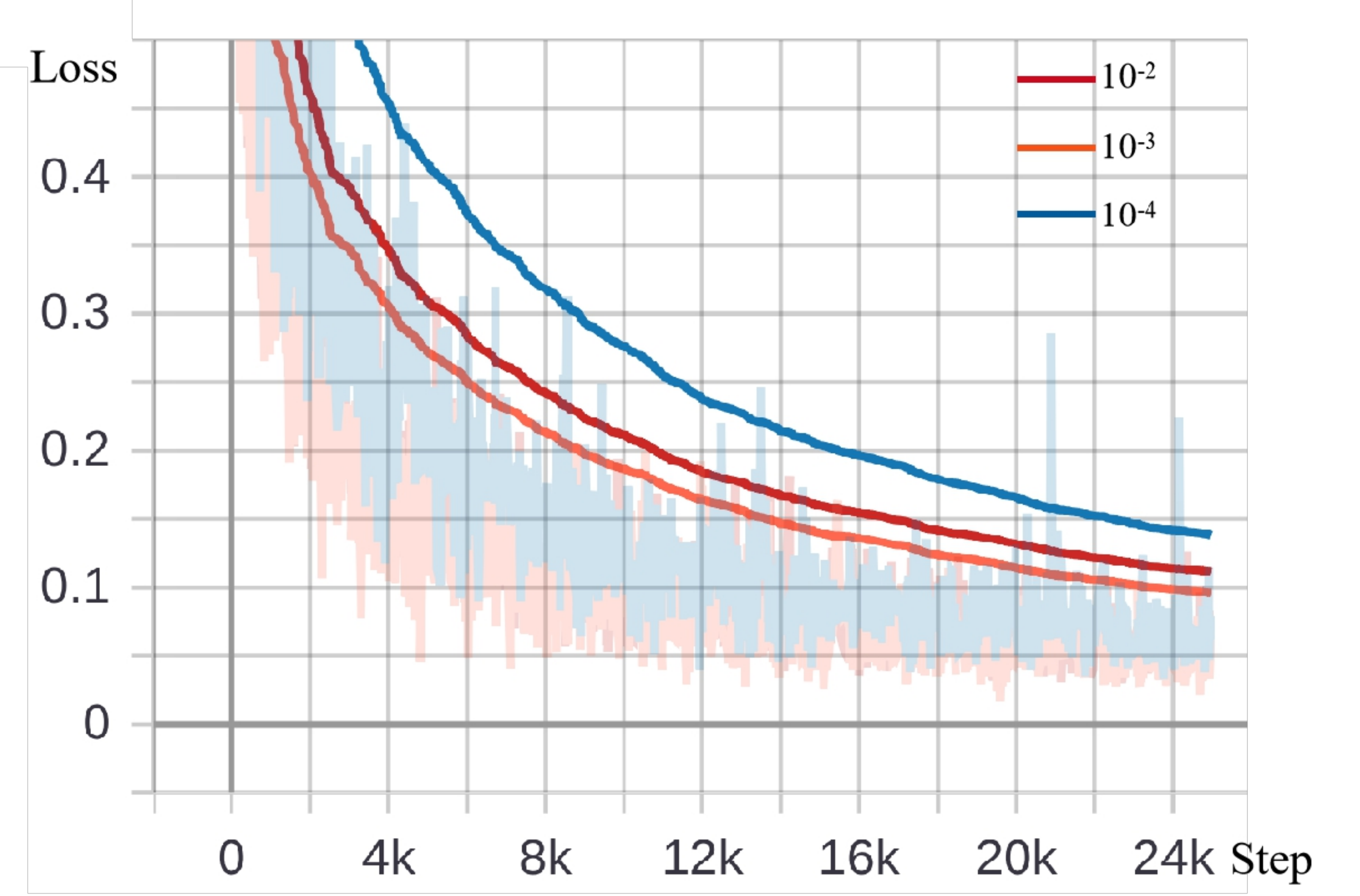}
\caption{The training loss of different learning rate.}
\label{fig:Dif_Learning_Rate}
\end{figure}

% 在预训练过程中,我们设置 50 个 epoch,每个 epoch 训练 500 次,并将学习率依次设置为 10 -2 ,10 -3 ,10 -4 对网络进行调整。通过不断地迭代 25000 次训练,实验结果如图1所示。
% 实验表明基于MAML的点云语义分割网络在训练过程中逐渐收敛。其中学习率太大活太小都会导致模型的收敛速度降低。因此本文选择最适合模型的学习率，即10-3.
In the pre-training process, we set 50 epochs, each epoch is trained 500 steps, and the learning rate is set to $10^{-2}$, $10^{-3}$, $10^{-4}$ in order. The experimental results are shown in Figure \ref{fig:Dif_Learning_Rate} after 25000 steps. Experiment shows that our network converges gradually during the training process. At the same time, the learning rate that is too large or too small will slow down the convergence of the model. Therefore, in this paper, we choose the most suitable learning rate for the model, which is $10^{-3}$.

% 为了更好的表示实验结果，我们采用了三种度量方法：总体准确率、平均类别准确率和平均交并比。它们的计算公式如表1所示。
For the purpose of better representing the experimental results, we used three metrics: overall accuracy (oAcc), mean class accuracy (mAcc) and mean class Intersection-over-Union (mIoU). Their calculation formulas are shown in Table \ref{tab:summary_statistical_features}.

\begin{table}[htb]
\caption{Summary of statistical features}
\begin{threeparttable}
\label{tab:summary_statistical_features}
\centering
\begin{tabular}{cll}
\toprule
No. & Name & Equation\tnote{*} \\
\midrule
1 & overall accuracy (oAcc)        & $\frac{\sum_{i=0}^{m}{c_{i}}}{\sum_{i=0}^{m}{n_{i}}}$ \\
\specialrule{0em}{3pt}{3pt}
2 & mean class accuracy (mAcc)    & $\frac {1}{m} \times \sum_{i=1}^{m}{\frac{c_{i}}{n_{i}}}$ \\
\specialrule{0em}{3pt}{3pt}
3 & mean class Intersection-over-Union (mIoU) & $\frac {1}{m} \times \sum_{i=1}^{m}{\frac{c_{i}}{n_{i}+w_{i}}}$ \\
\bottomrule
\end{tabular}
\begin{tablenotes}
        \footnotesize
        \item[*]  $m$ denotes the number of categories of data.\\ For the i-th category, the total number of point clouds is $n_{i}$, the number of correctly predicted point clouds is $c_{i}$, and the number of incorrectly predicted point clouds is $w_{i}$.
      \end{tablenotes}
\end{threeparttable}
\end{table}

\subsubsection{N-way K-shot random sampling}
\label{sec:experiment_exp1}
% 我们采用了n-way k-shot 的随机采样方法来构建应用于MAML算法的三维数据集。n和k的选取依次为2和6。以Area1数据集作为预训练数据，其他五个区域作为迁移应用的测试数据。实验结果如表所示。实验表明MAML算法能够成功地应用于三维点云环境。同时，不同数量的样本，模型的效果不同。当k相同时，2-way 的测试准确率比 6-way 的准确率高;当n相同时,6-shot 的测试准确率比 2-shot 的准确率高。因此本章实验选择2-way 6-shot进行相关实验。
We used the N-way K-shot random sampling method to construct the 3D dataset applied to the MAML algorithm. The parameters N and K were selected as 2 and 6, respectively. The Area 1 dataset was used as the pre-training data, and the other five areas were used as the test data for transfer learning. The experimental results are shown in Table \ref{tab:Area1-pre result}. The experiment shows that the MAML algorithm can be successfully applied to the 3D point cloud. At the same time, the model perform differently due to the different sizes of pre-training dataset. When K is the same, the accuracy of 2-way test is higher than that of 6-way test; when N is the same, the accuracy of 6-shot test is higher than that of 2-shot test. Therefore, the 2-way 6-shot method is chosen for the experiments in this section.

\begin{table}[H]
\caption{Semantic segmentation oAcc results for the S3DIS dataset after pre-training in Area 1}
\label{tab:Area1-pre result}
\centering
\begin{tabular}{ccccc}
\toprule
\multirow{2}{*}{Test dataset} & 
\multicolumn{2}{c}{2-way} & 
\multicolumn{2}{c}{6-way}\\
\cline{2-5}
~ & 2-shot & 6-shot & 2-shot & 6-shot \\
\midrule
Area 2 & 79.7 & 81.4 & 77.5 & 80.5 \\
Area 3 & 71.2 & 75.8 & 69.8 & 73.6 \\
Area 4 & 73.5 & 77.4 & 70.6 & 75.0 \\
Area 5 & 83.7 & 86.2 & 80.8 & 85.5 \\
Area 6 & 79.9 & 81.2 & 76.7 & 80.9 \\
\bottomrule
\end{tabular}
\end{table}

\subsubsection{Cross-validation}
\label{sec:experiment_exp2}
% 为了验证模型能够快速适应室内环境的不同场景变化，同时验证模型具有良好的泛化性能，我们进行了交叉验证实验。
% 在交叉验证实验过程中，我们分别使用6个区域作为预训练集对模型进行了训练，之后迁移到其他测试集上。语义分割结果如表所示。实验均使用了2-way 6-shot的任务采样方法。根据表格可以看到，使用Area3进行预训练并应用到Area1时总体准确率达到最低69.0%。而使用Area5进行预训练并应用到Area3时总体准确率高达88.0%。同时，使用Area5作为预训练集的总体准确率都高于其他预训练集。
% 表和表表明，预训练集数据量越大，所包含的语义类别越丰富。模型的效果也更好。同时测试集与预训练集相似性越高，测试效果越好。
To verify that our model can quickly adapt to changes in the indoor environment, and to verify that the model has good generalization performance, we conducted cross-validation experiments. During the cross-validation experiments, we trained the model using each of the six areas as a pre-training dataset, and later transferred it to the other test dataset. The semantic segmentation results are shown in Table \ref{tab:Cross_Val result}. The experiments all used the 2-way 6-shot sampling method. According to the Table \ref{tab:Cross_Val result}, the overall accuracy reached a minimum of 69.0$\%$ when Area 3 was used for pre-training and applied to Area 1, and reached a maximum of 88.0$\%$ when Area 5 was used for pre-training and applied to Area 3. Moreover, the overall accuracies using Area 5 as the pre-training dataset were higher than the other pre-training datasets. 

Table \ref{tab:Area1-pre result} and Table \ref{tab:Cross_Val result} show that the larger the amount of data in the pre-training dataset, the richer the semantic information included. And the model also works better. At the same time, if the test dataset has high similarity with the pre-trainging dataset, the model will perform better. 

\begin{table}[htb]
\caption{Semantic segmentation oAcc results for Cross-validation experiment}
\label{tab:Cross_Val result}
\centering
\begin{tabular}{ccccccc}
\toprule
% \diagbox{Training dataset}{Test dataset} & Area 1 & Area 2 & Area 3 & Area 4 & Area 5 & Area 6 \\
Test dataset & Area 1 & Area 2 & Area 3 & Area 4 & Area 5 & Area 6 \\
\midrule
Area 1 & —    & 81.4 & 75.8 & 77.4 & 86.2 & 81.2 \\
Area 2 & 71.1 & —    & 81.2 & 76.4 & 86.0 & 82.0 \\
Area 3 & \textbf{69.0} & 75.3 & —    & 70.8 & 79.7 & 74.5 \\
Area 4 & 75.4 & 78.5 & 74.7 & —    & 80.5 & 76.7 \\
Area 5 & \textbf{79.2} & \textbf{82.2} & \textbf{88.0} & \textbf{77.8} & \textbf{—}    & \textbf{84.7} \\
Area 6 & 77.0 & 80.3 & 85.8 & 78.6 & 83.1 & —    \\
\bottomrule
\end{tabular}
\end{table}

\subsubsection{Comparison of the effect of different test methods}
\label{sec:experiment_exp3}
\begin{figure*}[ht]
\centering
\includegraphics[width=5.0in]{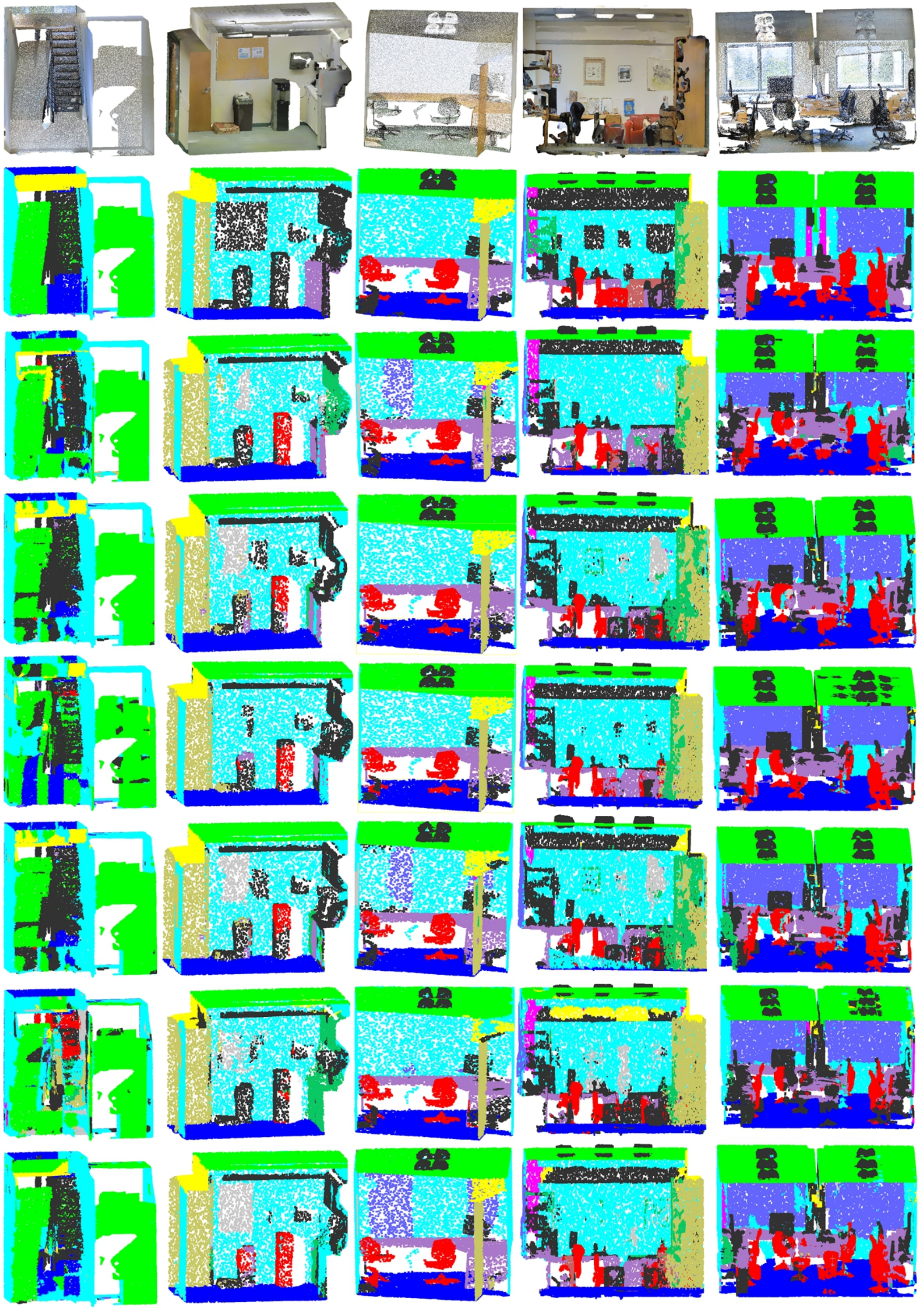}
\caption{Semantic segmentation results of five samples in Area 6. From left to right are hallway$\_$6, pantry$\_$1, office$\_$18, office$\_$7 and office$\_$8. From top to bottom are original point cloud, real semantic segmentation result, Area 1 as pre-training dataset, Area 2 as pre-training dataset, Area 3 as pre-training dataset, Area 4 as pre-training dataset, Area 5 as pre-training dataset and Area 1-5 as pre-training dataset.}
%\caption{Area 6 semantic segmentation results.}
\label{fig:Area6_test_Res}
\end{figure*}

% 我们以其他五个场景作为预训练集，以Area6作为迁移场景，语义分割准确率达到了87.8%。这种方法比其他单个场景作为预训练集的效果更好。结果进一步证明了预训练集包含的类别越丰富，模型的准确率越高。同时，模型具有更好的泛化能力。我们从Area6测试集中随机选择了5个测试样本，将其分割结果进行了可视化，如图。从左到右分别为hallway_6、pantry_1、office_18、office7 和 office_8。从上到下依次为原始点云、真实语义分割结果、Area1作为预训练集、Area2作为预训练集、Area3 作为预训练集、 Area4 作为预训练集、 Area5 作为预训练集、Area1-5作为预训练集。从每个房间来看，房间语义类别越复杂，分割效果越好，比如office_7 和 office_8,而语义类别越简单,分割效果反而不太好,比如 pantry_1。从总体上来看,使用 Area1-5 作为预训练集的语义分割结果比其他的好。
We used the other five areas as the pre-training dataset and Area 6 as the test dataset, and the semantic segmentation accuracy reached 87.8$\%$. This approach works better than other individual areas as a pre-training dataset. The results further demonstrate that the richer the categories contained in the pre-training dataset, the higher the accuracy of the model. Meanwhile, the model has better generalization ability. We randomly selected five test samples from the Area 6 test dataset and visualized their segmentation results as shown in Fig. \ref{fig:Area6_test_Res}. From left to right are hallway$\_$6, pantry$\_$1, office$\_$18, office$\_$7 and office$\_$8. From top to bottom are original point cloud, real semantic segmentation result, Area 1 as pre-training dataset, Area 2 as pre-training dataset, Area 3 as pre-training dataset, Area 4 as pre-training dataset, Area 5 as pre-training dataset, Area 1-5 as pre-training dataset. From the Fig. \ref{fig:Area6_test_Res}, we can see that the room segmentation results in complex semantic environment are better, such as office$\_$7 and office$\_$8, while the room segmentation results in simple semantic environment are not so good, such as pantry$\_$1. In general, the semantic segmentation results using Area 1-5 as the pre-training dataset are better than the others. 

\subsubsection{Comparison of the effect of different models}
\label{sec:experiment_exp4}

\section{Conclusion}
\label{sec:conclusion}

In our work, we propose a semantic segmentation method for point cloud based on meta-learning. Our method demonstrates that the MAML algorithm can be successfully applied to 3D point cloud. At the same time, with fewer samples, our network can be quickly adapted to new environments. Our network has good generalisation properties and can be used in a wide range of scenarios. It has great application value.

% or


% \appendices

% \section{}


\section*{Acknowledgment}
We sincerely thank the University of Electronic Science and Technology of China (UESTC) and Harbin Institute of Technology (HIT) for their technical support to this project.

\ifCLASSOPTIONcaptionsoff
  \newpage
\fi



%\bibliographystyle{IEEEtran}

% \bibliography{MyBib.bin}

\bibliographystyle{IEEEtran}
%\bibliography{IEEEfull,Bib/MyBib}
%\bibliography{Transactions-Bibliography/IEEEabrv,Bib/MyBib}
\bibliography{IEEEabrv,Few-Shot-Meta-Learning.bib}


% \end{thebibliography}


%\end{IEEEbiography}


%\newpage



%\vfill


\end{document}